\documentclass[sigconf]{acmart}
\AtBeginDocument{%
  }

\copyrightyear{2025}
\acmYear{2025}
\setcopyright{cc}
\setcctype{by}
\acmConference[MM '25]{Proceedings of the 33rd ACM International Conference on Multimedia}{October 27--31, 2025}{Dublin, Ireland}
\acmBooktitle{Proceedings of the 33rd ACM International Conference on Multimedia (MM '25), October 27--31, 2025, Dublin, Ireland}
\acmDOI{10.1145/3746027.3755596}
\acmISBN{979-8-4007-2035-2/2025/10}

\usepackage{latexsym}
\usepackage[T1]{fontenc}
\usepackage[utf8]{inputenc}
\usepackage{microtype}
\usepackage{graphicx}
\usepackage{amsmath}

\usepackage{amssymb}
\usepackage{array}
\usepackage{makecell}
\usepackage{float}
\usepackage{hyperref}
\usepackage{url}
\PassOptionsToPackage{prologue,dvipsnames, x11names, table}{xcolor}
\usepackage[dvipsnames]{xcolor}
\usepackage[x11names]{xcolor}
\usepackage[table]{xcolor}
\usepackage{xcolor}
\usepackage[most]{tcolorbox}
\usepackage{colortbl}
\usepackage{bbding}
\usepackage{pifont}
\usepackage{booktabs}
\usepackage{capt-of}
\usepackage{subfigure}
\usepackage{multirow}
\usepackage{longtable}
\usepackage{titletoc}
\usepackage{arydshln} 
\usepackage{enumitem}
\usepackage{balance}

\begin{document}

\title{Boosting Chart-to-Code Generation in MLLM via Dual Preference-Guided Refinement}

\author{Zhihan Zhang}
\affiliation{%
  \institution{Singapore Management University}
  \city{Singapore}
  \country{Singapore}}
\email{zhihanzhang.2024@phdcs.smu.edu.sg}

\author{Yixin Cao}
\affiliation{%
  \institution{Fudan University}
  \city{Shanghai}
  \country{China}}
\email{caoyixin2011@gmail.com}

\author{Lizi Liao}
\affiliation{%
  \institution{Singapore Management University}
  \city{Singapore}
  \country{Singapore}}
\email{lzliao@ smu.edu.sg}


\begin{abstract}
Translating chart images into executable plotting scripts—referred to as the chart-to-code generation task—requires Multimodal Large Language Models (MLLMs) to perform fine-grained visual parsing, precise code synthesis, and robust cross-modal reasoning. However, this task is inherently under-constrained: multiple valid code implementations can produce the same visual chart, and evaluation must consider both code correctness and visual fidelity across diverse dimensions. This makes it difficult to learn accurate and generalizable mappings through standard supervised fine-tuning.
To address these challenges, we propose a dual preference-guided refinement framework that combines a feedback-driven, dual-modality reward mechanism with iterative preference learning. Our approach introduces a structured variant generation strategy and a visual reward model to efficiently produce high-quality, aspect-aware preference pairs—making preference collection scalable and supervision more targeted. These preferences are used in an offline reinforcement learning setup to optimize the model toward multi-dimensional fidelity.
Experimental results show that our framework significantly enhances the performance of general-purpose open-source MLLMs, enabling them to generate high-quality plotting code that rivals specialized chart-centric models and even some proprietary systems. The code and datasets are publicly available at https://github.com/Zhihan72/Chart2Code.
\end{abstract}



\maketitle

\section{Introduction}

Charts serve as an essential medium for conveying structured information through visual representations, incorporating diverse visual elements such as colors, textual annotations, legends, and multi-panel subplots. While charts are widely used across scientific and analytical domains, understanding and reasoning over them remains a significant challenge in multimodal research \citep{masry-etal-2022-chartqa}. Recent advancements in Multimodal Large Language Models (MLLMs) have demonstrated remarkable capabilities in addressing a wide range of chart tasks, including chart question answering \citep{methani2020plotqa} and chart-to-text generation \citep{kantharaj-etal-2022-chart, kantharaj-etal-2022-opencqa, tang-etal-2023-vistext}. However, these tasks typically focus on high-level semantic understanding while overlooking the intricate visual structures embedded in charts, thereby limiting their evaluation depth and applicability. In response, the chart-to-code generation task has emerged \citep{shi2024chartmimicevaluatinglmmscrossmodal, yang-etal-2024-matplotagent}, which requires MLLMs to jointly perform fine-grained visual parsing, accurate code synthesis, and robust cross-modal reasoning from a chart image.

\begin{figure}[t]
    \centering
    \includegraphics[width=\linewidth]{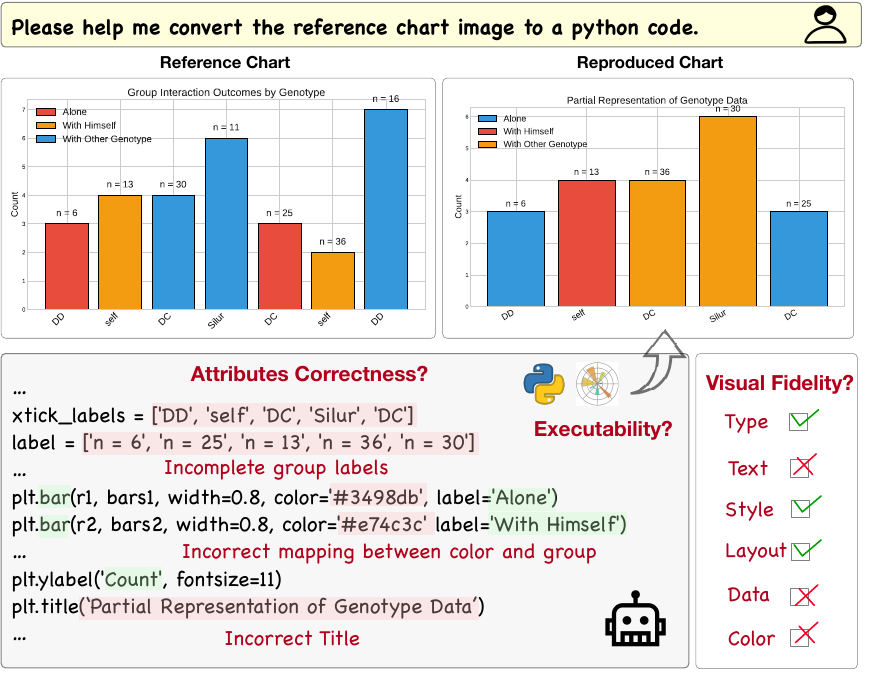}
    \vspace{-0.4cm}
    \caption{An illustrative example of the chart-to-code generation task, evaluated in dual-modality across multiple aspects.}
    \vspace{-0.2cm}
    \label{fig:task_case}
\end{figure}

The chart-to-code generation task is inherently under-constrained, presenting unique challenges for MLLMs to learn accurate and generalizable transformations from visual inputs to executable code \citep{tao2024codelutraboostingllmcode}. 
First, a plotting code and its rendered chart do not follow a one-to-one correspondence—multiple functionally correct implementations can produce the same chart while differing in syntax, plotting logic, or visual configuration. This inherent ambiguity limits the effectiveness of standard supervised fine-tuning (SFT), which relies on exact matches to a single reference and fails to account for the diversity of valid outputs across both code and visual modalities.
Second, chart construction depends on a complex combination of visual aspects—including chart type, color, text, layout, style, and underlying data—that together determine the chart’s appearance and semantics. This diversity further complicates the learning objective, as models must capture subtle variations across multiple dimensions to produce faithful code. 
Recent studies have shown that existing open-source MLLMs often generate incorrect or non-executable code with limited alignment to the input chart \citep{zhao2025chartcoderadvancingmultimodallarge}. These findings highlight the urgent need for an effective training paradigm that can align MLLMs with the specific demands of chart-to-code generation. 

To tackle these challenges, we propose \textbf{Chart2Code}, a dual preference-guided refinement framework designed to better align the training objective with the ultimate goal of chart-to-code generation task: \textit{generating executable code that faithfully reproduces the target chart}.
The framework is built upon two key components: a \textit{dual rewarding mechanism} and an \textit{iterative preference learning method}. The \textit{dual rewarding mechanism} provides fine-grained supervision by evaluating model-generated outputs across both code and image modalities. 
The code-side evaluation assesses the structural integrity and semantic correctness of the generated plotting script, while the image-side evaluation focuses on visual fidelity, measuring how well the rendered chart preserves the layout, styling, and perceptual attributes of the reference visualization.
Leveraging this dual-modality feedback, we introduce an \textit{iterative preference learning method} that progressively improves the model's performance. This offline Reinforcement Learning (RL) paradigm allows the evaluation of model-generated outputs against external synthetic codes through dual reward signals. The resulting preference pairs are then used to fine-tune the model via the Direct Preference Optimization (DPO) objective~\citep{rafailov2023direct}. At the end of each iteration, the updated model is evaluated on a new batch of reference charts for the subsequent iteration, enabling continuous refinement through dual-feedback-driven optimization. 

To enhance the effectiveness of preference learning, we develop a structured variant generation strategy and implement a visual reward model trained on a fine-grained, aspect-level feedback dataset during preference construction. The variant generation process produces synthetic code samples with controlled deviations from the gold-standard, enabling the creation of preference pairs that span varying levels of reproduction. To support accurate and interpretable image-side evaluation, we construct a feedback dataset comprising aspect-specific explanations and scores, which serve as supervision for training the visual reward model. This model enables reliable scoring of visual outputs, reinforcing the preference learning framework with fine-grained, aspect-aware guidance.

We validate our Chart2Code framework on three base MLLMs across two benchmarks and multiple evaluation metrics. Results demonstrate that our framework consistently yields substantial performance improvements under varying initialization settings, effectively aligning model outputs with the goal of generating high-quality, visually faithful plotting code—achieving performance comparable to specialized chart-centric models and even some proprietary systems. Our \textbf{main contributions} are:

\begin{itemize}[leftmargin=*]
\setlength\itemsep{0.2em}
\item We propose a dual preference-guided refinement framework (Chart2Code) that aligns MLLMs to the chart-to-code task via iterative preference learning.
\item We design a structured variant generation method and train a visual reward model, enabling high-quality, aspect-aware preference supervision across code and image modalities.
\item We achieve performance gains across multiple MLLMs and benchmarks, showing that our method boosts general-purpose models to match or surpass chart-specific and proprietary systems.
\end{itemize}

\section{Related Works}

\vspace{+0.1cm}\noindent \textbf{Multimodal Large Language Models.} MLLMs have emerged as a transformative paradigm in artificial intelligence, enabling joint reasoning over visual and textual inputs for a wide range of cross-modal tasks \citep{Ceylan2023Pix2VideoVE,yao2024minicpmvgpt4vlevelmllm}. Building upon the success of large language models (LLMs), recent efforts have focused on aligning visual and textual representations within a shared embedding space to facilitate effective multimodal understanding \citep{10.5555/3666122.3668264,yang2022zeroshot}. In parallel, there has been increasing interest in extending LLMs with multimodal instruction-following capabilities, allowing them to generate contextually grounded responses conditioned on both visual content and textual prompts \citep{Ye2023mPLUGOwlME, zhu2024minigpt}. 

\vspace{+0.1cm}\noindent \textbf{Preference Learning.} 
Preference learning has emerged as a prominent approach to enhance the performance of LLMs by aligning them with human preferences, serving as the foundation of RL from Human Feedback (RLHF) \citep{10.5555/3600270.3602281, casper2023open, rafailov2023direct}. Recent advancements of offline preference optimization techniques such as DPO \citep{rafailov2023direct}, are becoming more popular for their simplicity and efficiency. Iterative variants of these offline methods have demonstrated effectiveness in progressively refining model outputs through repeated optimization over newly constructed preference pairs \citep{adolphs-etal-2023-cringe, Xiong2023GibbsSF, pang2024iterative}. Although RLHF has been extensively explored in LLMs, its adaptation to MLLMs remains relatively underexplored. Existing approaches for MLLM alignment typically construct preference datasets either by leveraging externally annotated synthetic examples \citep{li-etal-2024-vlfeedback, zhou2024aligning, sun-etal-2024-aligning} or by employing self-sampling with reward-based ranking \citep{deng2024enhancing, zhou2024calibrated, 10.5555/3692070.3692326, 10.5555/3692070.3694459}. In the chart-to-code generation setting, we combine model-generated codes with synthetic variants for iterative preference learning, exploring how this hybrid approach enhances model performance.

\vspace{+0.1cm}\noindent \textbf{Chart-to-code Generation.} The chart-to-code generation task has recently attracted growing attention in the research community \citep{yang-etal-2024-matplotagent, he2024distillvisualchartreasoning, zhao2025chartcoderadvancingmultimodallarge}, which requires models to synthesize executable code grounded in fine-grained visual understanding. Prior work has focused on constructing benchmarks through large-scale chart collection and human annotation to evaluate model performance in this setting \citep{shi2024chartmimicevaluatinglmmscrossmodal, wu2024plot2codecomprehensivebenchmarkevaluating}. While several studies have explored enhancing the MLLM's chart-to-code generation capability via SFT on curated datasets \citep{zhao2025chartcoderadvancingmultimodallarge, han2023chartllama, masry-etal-2024-chartinstruct}, our work is the first to introduce the offline RL paradigm to align MLLMs with the inherently dual-modality and multi-dimensional requirements of chart-to-code generation.

\begin{figure*}[t]
    \centering
    \includegraphics[width=\linewidth]{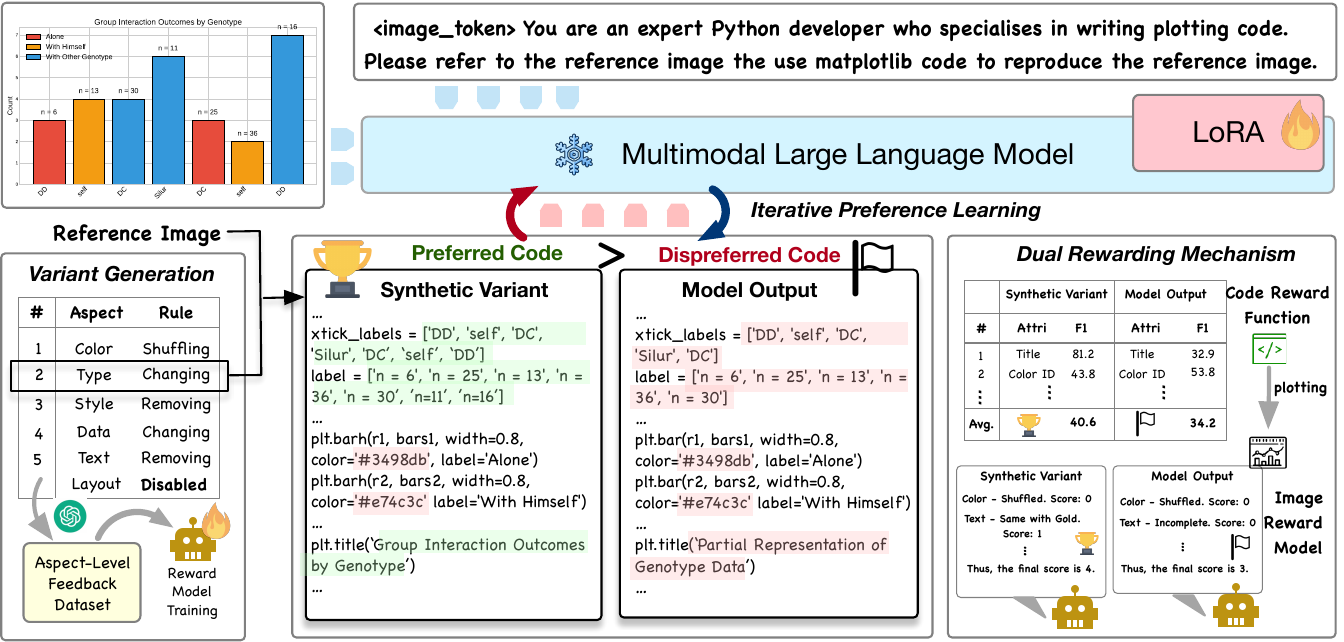}
    \caption{Overview of Chart2Code. It consists of two core components: a dual rewarding mechanism that provides fine-grained feedback via a heuristic F1-based code scorer and a visual reward model, and an iterative preference learning process that refines the model through DPO optimization. To enable high-quality preference construction, we introduce a structured variant generation strategy and an aspect-level feedback dataset for training the visual reward model in image-side evaluation.}
    \label{fig:pipeline_method}
\end{figure*}

\section{Our Method: Chart2Code}
\label{section:method}

We propose \textbf{Chart2Code}, a novel dual preference-guided refinement framework designed to enhance the chart-to-code generation capabilities of MLLMs by better aligning the training objective with the task’s inherently dual-modality and multi-dimensional nature. The framework comprises two core components: 1) \textit{dual-modality and fine-grained rewarding mechanism}, which delivers fine-grained feedback from both image and code perspectives, guiding the model’s output refinement through multi-dimensional evaluation; and 2) \textit{iterative preference learning process}, which adopts an offline RL paradigm to collect dual-modality feedback on model outputs and progressively refine the model in subsequent iterations.

\subsection{Problem Setting}

Given a chart image  $I_i^g$  and an instruction  $x_i$  ( $i \in [1,N]$ ), where  $g$  denotes the gold-standard reference and  $N$  represents the size of the gold code dataset, the target model  $M_{t}$  is tasked with generating code  $C_i^0$  to replicate the reference image, where $t$ is an integer to indicate the current iteration ($t \in [0,T]$). Formally, 
$$C_i^0 = M_{t}(I_i^g, x_i).$$
This chart-to-code generation task is typically approached with supervised fine-tuning (SFT), where models are trained to mimic gold-standard scripts $C_i^g$ \citep{zhao2025chartcoderadvancingmultimodallarge, chen2024onechart, meng-etal-2024-chartassistant}. However, due to the under-constrained nature of the task—where multiple valid code implementations can yield visually similar charts—SFT often fails to provide sufficiently flexible supervision needed to generalize beyond the reference outputs. 

Motivated by recent advancements in RL and preference-based optimization \citep{rafailov2023direct, deng2024enhancing}, we adopt a preference learning framework that allows the model to learn from relative comparisons between diverse outputs. By incorporating fine-grained, dual-modality reward signals $r_i$—evaluating both the generated code $C_i^0$ and its rendered image $I_i^0$—this approach aligns the training objective more closely with the true goal: \textit{generating executable code that faithfully reproduces the target chart}.

\subsection{Dual Rewarding Mechanism}
\label{section:reward_modeling}

\subsubsection{Dual Modality} 
We propose a dual rewarding mechanism that provides robust and comprehensive supervision by jointly evaluating both the generated code $C_i^0$ and its corresponding rendered image $I_i^0$. This mechanism produces a code-side reward $r_i^C$ and an image-side reward $r_i^I$, each reflecting different but complementary aspects of output quality. The code-side evaluation focuses on the internal structure and semantic correctness of the generated script. It verifies whether the code uses appropriate plotting APIs, encodes data relationships correctly, and adheres to syntax and logic constraints. In contrast, the image-side evaluation emphasizes external fidelity to the reference visualization, assessing how accurately the rendered chart reproduces the intended layout, styling, and perceptual attributes. This includes aspects such as spatial arrangement of subplots, font size and position of text labels, color mapping, and visual balance—features that may not be explicitly captured in the code structure but are critical to the visual appearance of the chart.

Using this dual reward mechanism, we construct high-quality preference pairs to guide the preference learning process. Specifically, we compare two generated outputs $C_i^x$ and $C_i^y$ ($x \neq y$) and retain the pair only if one sample strictly outperforms the other one in both reward modalities—that is, $r_{i,x}^C > r_{i,y}^C$ and $r_{i,x}^I > r_{i,y}^I$. This requirement ensures that preference labels reflect consistent superiority across both the code and image perspectives.

\subsubsection{Fine-grained Reward}
\label{section:fine_grained_reward}
We introduce the rewarding methods for both the code and image modalities to enable fine-grained supervision during preference learning. On the code side, we first execute each code sample $C_i^j$ to ensure it uses appropriate plotting APIs and adheres to basic syntactic and logical constraints. If execution fails, the sample is assigned a reward score of zero in both modalities. For executable samples, we compute the code-side reward using the heuristic F1-based scoring method \citep{shi2024chartmimicevaluatinglmmscrossmodal}. This lightweight approach avoids complex code reasoning by tracing key semantic attributes during execution—such as color identifiers, titles, and data tables—which are individually compared to their counterparts in the gold-standard code using F1 score computation. The final code-side reward $r_{i,j}^{C}$ is calculated as the average of these attribute-level F1 scores with a range of 0–100.

On the image side, we evaluate the visual fidelity of the rendered chart $I_i^j$ by comparing it to the reference image $I_i^g$ using a multi-aspect binary scoring method. This method assigns a binary sub-score (0 or 1) to each of six predefined visual aspects: \textit{chart type}, \textit{data}, \textit{layout}, \textit{color}, \textit{text}, and \textit{style}. Each sub-score reflects whether the corresponding aspect in the generated chart aligns with the reference, considering all relevant visual cues. The total image-side reward is computed as the sum of aspect-specific sub-scores:
$r_{i,j}^I = \sum_{k=1}^{6} r_{i,j,k}^I,\quad r_{i,j,k}^I \in {0, 1}.$
To automate this fine-grained evaluation, we train a visual reward model $M_e$ to predict the aspect-level sub-scores based on visual differences between the generated and reference charts. The model is trained on an aspect-level feedback dataset, as described in Section~\ref{section:feedback_collection}.

\subsection{Iterative Preference Learning}
\label{section:iterative_preference}

We adopt an offline RL paradigm to iteratively guide the model toward generating executable code that faithfully reproduces the target chart. In each iteration, we begin with a set of gold-standard (code, image, instruction) triplets, denoted as $\mathcal{D}_t^g = \{(C_i^g, I_i^g, x_i)\}_{i=1}^{N}$, along with the current model checkpoint from the previous iteration, $M_t$. Based on generated outputs, we construct a dataset of generated samples $\mathcal{D}_t^v = \{D_i^v\}_{i=1}^{N}$, where each instance $D_i^v$ consists of a collection of code–image–reward tuples:
$$
D_i^v = \{(C_i^j, I_i^j, r_{i,j}^C, r_{i,j}^I) \mid 0 \leq j \leq k \},
$$
with each $C_i^j$ representing either a model-generated or synthetically perturbed code sample (detailed in Section~\ref{section:variant_gen}), and $k$ indicating the total number of samples associated with the $i$-th reference example. Each code sample is paired with its corresponding rendered chart image and evaluated using the dual rewarding mechanism, resulting in code-side ($r_{i,j}^C$) and image-side ($r_{i,j}^I$) reward scores.

To construct the preference dataset, we pairwise all combinations of code samples in $D_i^v$ and select the preferred sample in each pair based on their dual reward scores. This results in a preference-labeled dataset:
$$
D_i^p = \{(I_i^g, x_i, C_i^{w_m}, C_i^{l_m}) \mid 1 \leq m \leq \tfrac{n(n-1)}{2} \},
$$
where $C_i^{w_m}$ and $C_i^{l_m}$ denote the winning and losing code samples, respectively, and $n$ is the total number of samples in $D_i^v$. The number of possible preference pairs is upper-bounded by $n(n-1)/2$. To ensure the quality of supervision, we discard any pair where the two samples receive identical scores in either the code-side or image-side evaluation, thereby enforcing strict agreement across both modalities.

Using the preference pairs, we train a new model $M_\theta$ , leveraging the previous iteration’s model  $M_t$  as the reference model in the denominator of DPO loss function \citep{rafailov2023direct}. The model parameter $\theta$ is updated as follows:

\begin{equation*}
\begin{split}
\mathcal{L}_{DPO}(C_i^{w_m}, C_i^{l_m} &| I_i^g, x_i) = \\
& - \log \sigma \left( \beta \frac{M_{\theta}(C_i^{w_m}| I_i^g, x_i)}{M_{t}(C_i^{w_m}| I_i^g, x_i)} - \beta \frac{M_{\theta}(C_i^{l_m}| I_i^g, x_i)}{M_{t}(C_i^{l_m}| I_i^g, x_i)} \right),
\end{split}
\end{equation*}
where $\sigma$ is the sigmoid function. At the end of this training, we obtain the updated model $M_{t+1}=M_\theta$ , which is then used to generate data for the subsequent iteration. 

In practice, RLHF typically begins by fine-tuning a pre-trained model on high-quality, task-specific data using supervised learning, resulting in an initial model $M_0 = M_{\text{SFT}}$ \citep{rafailov2023direct, xiong2024llavacriticlearningevaluatemultimodal}. Following this paradigm, we train our model on a set of gold-standard samples $\mathcal{D}_0^g = \{(C_i^g, I_i^g, x_i)\}$ prior to initiating our Chart2Code framework, to mitigate distributional mismatch between the true reference distribution and the policy distribution used during DPO.

\section{Preference Construction}

To support effective preference learning, we develop a structured variant generation strategy that produces code variants with controlled levels of deviation from the gold-standard code. In parallel, we collect a feedback dataset containing detailed explanations and aspect-level annotations, which serves as supervision for training our visual reward model used in fine-grained image-side evaluation.

\subsection{Rule-based Variant Generation}
\label{section:variant_gen}

The rule-based variant generation strategy constructs a set of code variants that exhibit controlled levels of deviation from a given gold-standard script. Starting from a predefined set of visual aspects and their corresponding transformation rules, we sample a variation path and employ GPT-4o to generate code variants through progressive, aspect-level modifications. These synthetically perturbed variants, together with model-generated outputs, are used to construct preference pairs, effectively bridging the gap between ideal references and real model behavior.

\subsubsection{Aspects and Rules} We develop a structured variant generation strategy grounded in six well-defined aspects $A = \{a_k : k \in [1,6]\}$, capturing the full spectrum of differences that can arise between two charts.  Specifically, \textit{type} refers to the detailed chart format (e.g., donut pie chart, stacked bar chart); \textit{data} focuses on the structure and values of the underlying dataset; \textit{layout} captures the arrangement and number of subplots; \textit{color} evaluates the color schemes applied to different data groups; \textit{text} includes all textual elements such as axis labels and titles; and \textit{style} pertains to aesthetic properties such as grid lines, borders, and marker shapes.

To enable controlled perturbations along each aspect, we define a corresponding set of transformation rules $R_k = \{r_{j,k} : j \in [1, n_k]\}$ for each $a_k$. These rules operate on the reference code $C_i^g$ to generate diverse variants by selectively modifying, replacing, or removing relevant components. For example, \textit{type} transformations are guided by a predefined dictionary mapping each chart type to its alternatives; \textit{data} modifications involve deleting, altering, or fabricating data groups or dimensions; \textit{layout} changes include rearranging or omitting subplots; \textit{text} variations affect group labels, titles, and axis annotations through rewriting or removal; \textit{color} alterations involve shuffling color schemes or reducing color diversity; and \textit{style} adjustments toggle visual elements such as grids, borders, or legends. This structured and interpretable variation strategy allows us to generate aspect-specific variants that support consistent preference supervision and effective model training. The complete definition of aspects and rules are detailed in supplementary material.

\subsubsection{Variation Path Sampling}

Given the defined aspects and associated transformation rules, we sample a variation path for each gold-standard code, consisting of a sequence of aspect-rule pairs. 
Formally, given a reference code $C_i^g$, we first sample a sequence of distinct aspects $A_i = \{a_{i,\hat{k}} : 1 \leq \hat{k} \leq 6\}$, where each $a_{i,\hat{k}}$ is uniquely selected from the defined aspect set. For each selected aspect, we randomly sample a transformation rule $u_{i,\hat{k}} \in R_{a_{i,\hat{k}}}$, resulting in a variation path defined as $P_i^v = \{(a_{i,\hat{k}},u_{i,\hat{k}}) : 1 \leq \hat{k} \leq 6\}$. Certain aspects, such as \textit{type} and \textit{layout}, may be inapplicable depending on the chart (e.g., non-editable types or absence of multiple subplots), resulting in variation paths ranging from 4 to 6 steps in length. 

\subsubsection{Variant Generation}
\label{section:variant_generation}
For each gold-standard code $C_i^g$, we leverage GPT-4o \citep{gpt-4o-website} to generate structured code variants along two randomly sampled variation paths, following the self-instruct paradigm \citep{wang-etal-2023-self-instruct}. 
Given a variation path $P_i^v$, we iteratively apply each transformation rule in sequence: at step $\hat{k}$, GPT-4o is provided with the most recent variant $C_i^{\hat{k}-1}$, the selected rule $u_{i,\hat{k}}$, and an instruction prompt to generate the next variant $C_i^{\hat{k}}$. This perturbation process produces a set of progressively modified variants $\{C_i^{\hat{k}} : 1 \leq \hat{k} \leq 6\}$, where each variant differs from the original reference code by $\hat{k}$ aspect-level transformations. This structured generation strategy enables consistent, interpretable preference ranking across varying levels of visual fidelity. A complete example of the variation path, variant generation, and the prompt design is provided in the supplementary material.

\subsection{Aspect-level Feedback Collection}
\label{section:feedback_collection}

To train the visual reward model for our multi-aspect binary scoring mechanism (Section~\ref{section:fine_grained_reward}), we construct an aspect-level feedback dataset derived from the transformation history of code variants along sampled variation paths \citep{xiong2024llavacriticlearningevaluatemultimodal}. This dataset captures detailed explanations of how each variant diverges from its reference across specific visual aspects, enabling the reward model to learn fine-grained, aspect-aware reasoning for evaluating visual fidelity in a structured and interpretable manner.

\subsubsection{Feedback Composition}
Each training instance in the feedback dataset is structured as: (\texttt{Reference Image}, \texttt{Task Instruction}, \texttt{Generated Image}, \texttt{Evaluation Criteria}, \textcolor{olive}{\texttt{Score}}, \textcolor{olive}{\texttt{Explanation}}). \texttt{Evaluation Criteria} prompts the reward model to assign a binary sub-score to each of the six predefined visual aspects of the \texttt{Generated Image}, compared with \texttt{Reference Image}. The highlighted outputs, \textcolor{olive}{\texttt{Score}} and \textcolor{olive}{\texttt{Explanation}}, correspond to the reward model’s expected predictions: six binary sub-scores and detailed, aspect-level explanation for each decision, forming the basis of supervised training for the visual reward model.

\subsubsection{Feedback Collection} 

To construct training data for the reward model, we extract aspect-level scores and explanations from the transformation history of code variants generated along sampled variation paths by GPT-4o (Section~\ref{section:variant_generation}). For example, consider a variant located at the third step of a variation path, where the modified aspects include layout (step 1), text (step 2), and style (step 3). We retrieve the corresponding explanations for layout and text from their respective transformation steps, and collect the explanation for style at the current step. Each of these deviated aspects is assigned a binary score of 0 to reflect a mismatch with the reference image. The rest aspects, which remain unaltered, are assigned a score of 1 and paired with a standardized explanation: The response meets the requirements in this aspect. To ensure consistency and clarity across the dataset, each feedback instance is further refined using GPT-4o to produce well-structured, aspect-specific justifications in a unified format. A complete example of feedback collection is detailed in supplementary material.

\begin{table}
  \caption{Distribution of numbers of gold codes, variants, and preference (Pref.) pairs across iterations.}\vspace{-0.1cm}
  \label{table:data_stat}
  \scalebox{0.9}{
  \begin{tabular}{lccc}
    \toprule
    Phrase & Gold Code & Variant & Pref. Pair \\ 
    \midrule
    Iteration 1 & 300 & 2,752 & 7,802 \\ 
    Iteration 2 & 300 & 2,710 & 7,680\\ 
    Iteration 3 & 300 & 2,694 & 7,590 \\ 
    \hline
    \textit{Total} & 900 & 11,906 & 23,072 \\
  \bottomrule
\end{tabular}}
\end{table}

\begin{table*}[t]
    \centering
    \caption{Performance of baselines and trained models across iterations in the Chart2Code framework on two chart-to-code datasets. The bold content represents the highest value in the category. }\vspace{-0.1cm}
    \label{table:experimental_result}
    \setlength\tabcolsep{2pt}
    \scalebox{0.95}{
    \begin{tabular}{p{0.2cm}p{2.8cm}p{1.5cm}<{\centering}p{1.5cm}<{\centering}p{1.5cm}<{\centering}p{1.5cm}<{\centering}p{0.2cm}<{\centering}p{1.5cm}<{\centering}p{1.5cm}<{\centering}p{1.5cm}<{\centering}p{1.5cm}<{\centering}}
        \toprule
          & &  \multicolumn{4}{c}{\textbf{ChartMimic}} & & \multicolumn{4}{c}{\textbf{Plot2Code}} \\
        \cmidrule(r){3-6} \cmidrule(r){8-11}
         Models & & Exec. Rate &  Heuristic F1 &  GPT Conti. & Multi- Binary & & Exec. Rate &  Heuristic F1 &  GPT Conti. & Multi- Binary \\
        \midrule
        & & \multicolumn{9}{c}{\textit{Propriety Multimodal Large Language Models}} \\
        \midrule
        \multicolumn{2}{l}{Gemini Pro Vision}  & 64.2  & 45.0  & 38.1  & 3.47 &  & 66.3 & 18.7 & 43.9 & 3.36\\
        \multicolumn{2}{l}{Claude-3-Opus} & 86.4  & 56.0  & 45.4  & 3.62 &  & \textbf{87.1} & 27.4  & 51.9 & \textbf{3.58}\\
        \multicolumn{2}{l}{GPT-4V}  & \textbf{91.4}  & \textbf{74.3}  & 68.4  & 3.87 &  & 86.9 & \textbf{31.4} & 57.1 & 3.28 \\
        \multicolumn{2}{l}{GPT-4o-mini} & 85.6  & 67.6  & \textbf{70.0}  &  \textbf{3.95}  &  & 79.8 & 28.3  & \textbf{58.6} & 3.41 \\
        \midrule
        & &\multicolumn{9}{c}{\textit{Chart-augmented Multimodal Large Language Models}} \\
        \midrule
        \multicolumn{2}{l}{ChartInstruct-7B} & 1.3 & 0.4 & 1.8 & 0.07  &  & 2.5 & 0.7 & 1.1 & 0.05 \\
        \multicolumn{2}{l}{ChartVLM-L-14B}  & 12.0 & 3.9 & 3.4 & 0.18 & & 15.9 & 2.0 & 2.3  & 0.15 \\
        \multicolumn{2}{l}{ChartLlama-13B}  & \textbf{55.4} & \textbf{11.7} & \textbf{12.6} & \textbf{0.47} &  & \textbf{80.3} & \textbf{14.5} & \textbf{24.8} & \textbf{1.39} \\
        \midrule
        & &\multicolumn{9}{c}{\textit{Open-source Multimodal Large Language Models}} \\
        \midrule
        \multicolumn{2}{l}{InternVL2.5-2B}  & 48.8  & 21.9  & 22.6  & 1.31 &  & 54.5 & 11.0  & 21.5 & 1.41 \\
        \multicolumn{2}{l}{InternVL2.5-8B}  & 54.2 & 23.4 & \textbf{32.1} & \textbf{1.78} & & \textbf{79.5} & \textbf{17.2} & \textbf{40.5} & \textbf{2.18} \\
        \multicolumn{2}{l}{Qwen2-VL-2B} & 60.9 & 28.7 & 29.6 & 1.01 & & 59.8 & \textbf{17.2} & 27.6 & 1.27 \\
        \multicolumn{2}{l}{Qwen2-VL-7B} & \textbf{62.2} & 30.0 & 28.9  & 1.09 &  & 61.4 & 17.1 & 26.4 & 1.69 \\
        \multicolumn{2}{l}{MiniCPM-Llama3-V2.5} & 58.2 & \textbf{30.2} & 24.2 & 1.21 & & 58.4 & 16.2 & 22.3 & 1.33 \\
        \multicolumn{2}{l}{LLaVA-v1.6-7B}  & 55.6 & 23.6  & 20.2 & 1.09 &  & 60.6 & 12.8 & 20.1 & 1.13  \\
        \hline
        \multicolumn{2}{l}{Chart2Code (LLaVA-v1.6-7B)}  &  &   &  &  &  &  &  &  &  \\
        & \textit{Initial 3k SFT ($M_0$)}  & 56.2  & 24.8  & 23.2  & 1.38 &  &  57.6 & 11.6  & 21.5 & 1.09 \\
        & Iteration 1 ($M_1$) & 58.2 & 26.9 & 24.6 & 1.46 & & 61.2 & 13.2 & 23.4 & 1.21 \\
        & Iteration 2 ($M_2$) & 62.2 & \textbf{27.3} & 25.5 & \textbf{1.53} & & 64.0 & 17.8 & 25.6 & 1.35 \\
        & Iteration 3 ($M_3$)  & \textbf{63.2} & 27.2 & \textbf{25.8} & 1.52 &  & \textbf{66.8} & \textbf{19.4} & \textbf{32.8} & \textbf{1.45} \\
        \cmidrule(r){2-11}
        & \textit{Initial 160k SFT ($M_0$)}  & 71.8  & 62.3 & 35.6 & 2.83 &  & 73.5 & 22.2  & 39.5 & 2.71 \\
        & Iteration 1 ($M_1$) & 77.2 & 63.1 & 38.4 & 3.15 & & 72.7 & 20.6 & 36.4 & 2.82  \\
        & Iteration 2 ($M_2$) & 79.6 & 65.3 & 41.0 & 3.09 & & 80.8 & 24.4 & 42.2 & 3.28 \\
        & Iteration 3 ($M_3$)   &  \textbf{84.6} & \textbf{69.2} & \textbf{42.1} & \textbf{3.38} &  & \textbf{83.3}  & \textbf{24.8}  & \textbf{48.5} & \textbf{3.64} \\
        \bottomrule
    \end{tabular}
    }
\end{table*}

\subsection{Dataset Statistics}

\subsubsection{Source Data}
\label{section:source_data}
Our data source consists of the plotting scripts of ReachQA training set (3,249) \citep{he2024distillvisualchartreasoning} and ChartCoder-160k \citep{zhao2025chartcoderadvancingmultimodallarge}, each serving as a different SFT initialization setting for our Chart2Code framework. Moreover, we employ the self-instruct method \citep{NEURIPS2020_1457c0d6} through GPT-4o to generate gold-standard code-image pairs $\mathcal{D}_t^g$ for each iteration. The detailed prompt is provided in the supplementary material.

\subsubsection{Preference Dataset Details}
\label{section:ref_dataset_details}
Our dataset comprises 11,906 variants, and 23,072 preference pairs (including model's outputs), with their distribution across three iterations summarized in Table~\ref{table:data_stat}. Notably, the feedback dataset consists of 3,750 instance, with 10\% reserved for evaluation. During code generation, non-executable codes are discarded, accounting for 3.9\% of the total (excluded from the above counts). For each gold code, we sample two variation paths, with a maximum path length of five. The proportion of paths involving each aspect is as follows: 97.9\% for \textit{data}, 97.2\% for \textit{text}, 97.7\% for \textit{style}, 97.8\% for \textit{color}, 56.0\% for \textit{type}, and 17.3\% for \textit{layout}. The relatively lower proportions for type and layout are due to the limited number of images that support type modifications or involve multiple subplots.

\section{Experiment}
We validate our Chart2Code framework on open-source MLLMs across two benchmarks and multiple evaluation metrics. To evaluate the robustness of our framework under different training conditions, we experiment with two SFT initialization settings using 3k and 160k training examples, respectively. Furthermore, we conduct a series of ablation studies to evaluate the framework’s generalizability across model architectures, as well as the individual contributions of the dual-modality reward mechanism and the preference construction strategy.

\subsection{Experimental Settings}
\subsubsection{Model and Baselines.}
We evaluate a diverse set of MLLMs across two categories: 
(1) Proprietary models, including GPT-4o \citep{gpt-4o-website}, GPT-4o-mini \citep{gpt-4o-mini-website}, Claude-3-Opus \citep{claude-3-website}, and Gemini Pro Vision \citep{geminiteam2024geminifamilyhighlycapable}. 
(2) Chart-augmented open-source models, such as ChartInstruct-7B \citep{masry-etal-2024-chartinstruct}, ChartLlama-13B \citep{han2023chartllama}, ChartVLM-L-14B \citep{xia2024chartx}, and ChartCoder-7B \citep{zhao2025chartcoderadvancingmultimodallarge}.
(3) Latest open-source models, including LLaVA-v1.6-7B (Mistral version) \citep{liu2023visual}, InternVL2.5-2B, InternVL2.5-8B \citep{chen2024internvl}, Qwen2-VL-2B, Qwen2-VL-7B \citep{Qwen2VL}, and MiniCPM-Llama3-V2.5 \citep{yao2024minicpmvgpt4vlevelmllm}.

We conduct our iterative training framework primarily on LLaVA-v1.6-7B, and further evaluate its effectiveness through ablation studies using InternVL2.5-2B and Qwen2-VL-7B. In addition, we employ Phi-3.5-Vision \citep{abdin2024phi3technicalreporthighly} as the backbone for training the visual reward model through SFT, leveraging its strong cross-modal reasoning capabilities and native support for multi-image input.

\subsubsection{Evaluation Datasets and Metrics.}
We evaluate our models on two widely used chart-to-code benchmarks: ChartMimic \citep{shi2024chartmimicevaluatinglmmscrossmodal} and Plot2Code \citep{wu2024plot2codecomprehensivebenchmarkevaluating}, containing 500 and 132 examples respectively. For evaluation metrics, we adopt the two metrics from our dual rewarding mechanism—namely, the heuristic F1-based code scoring (Heuristic F1) and the multi-dimensional binary scoring (Multi-Binary) for image fidelity. Additionally, we employ GPT-4o continuous scoring (GPT Conti.), which has been commonly used in recent works \citep{shi2024chartmimicevaluatinglmmscrossmodal, yang-etal-2024-matplotagent, zhao2025chartcoderadvancingmultimodallarge}. This method prompts GPT-4o to assess the similarity between the generated and reference chart images using an open-ended, perception-driven evaluation process, assigning a continuous score ranging from 0 to 100 without relying on strict criteria. The detailed prompts for evaluation are provided in the supplementary material.

\subsubsection{Implementation Details.} We perform the preference learning over one epoch per iteration and evaluate two SFT initialization settings using 3k and 160k training examples, respectively, to assess the generalizability of the Chart2Code framework under varying initializing conditions (Section~\ref{section:source_data}). All training runs adopt consistent LoRA fine-tuning hyperparameters \citep{hu2022lora}, with \texttt{lora\_r = 128} and \texttt{lora\_alpha = 256}. The learning rates are set to \texttt{2e-4} for SFT and \texttt{2e-5} for DPO optimization, with a global batch size of 8 for all experiments. A complete record of training environment, settings and procesures is provided in the supplementary material.

\subsection{Main Results}

As shown in Table~\ref{table:experimental_result}, our Chart2Code framework significantly and consistently improves the performance of the base MLLM across execution rate, code quality, and image fidelity under both 3k and 160k SFT initialization settings. These results demonstrate the robustness and effectiveness of our framework across varying initialization conditions. 
Notably, LLaVA-v1.6-7B achieves an impressive execution rate of 84.6\% under the 160k initialization setting—on par with GPT-4o-mini—while also delivering substantial gains in both code quality and visual fidelity across iterations. This demonstrates that Chart2Code not only improves executability but also more effectively guides models toward generating semantically richer and structurally accurate code—surpassing the limitations of standard supervised fine-tuning.

\begin{figure}
    \centering
    \includegraphics[width=\linewidth]{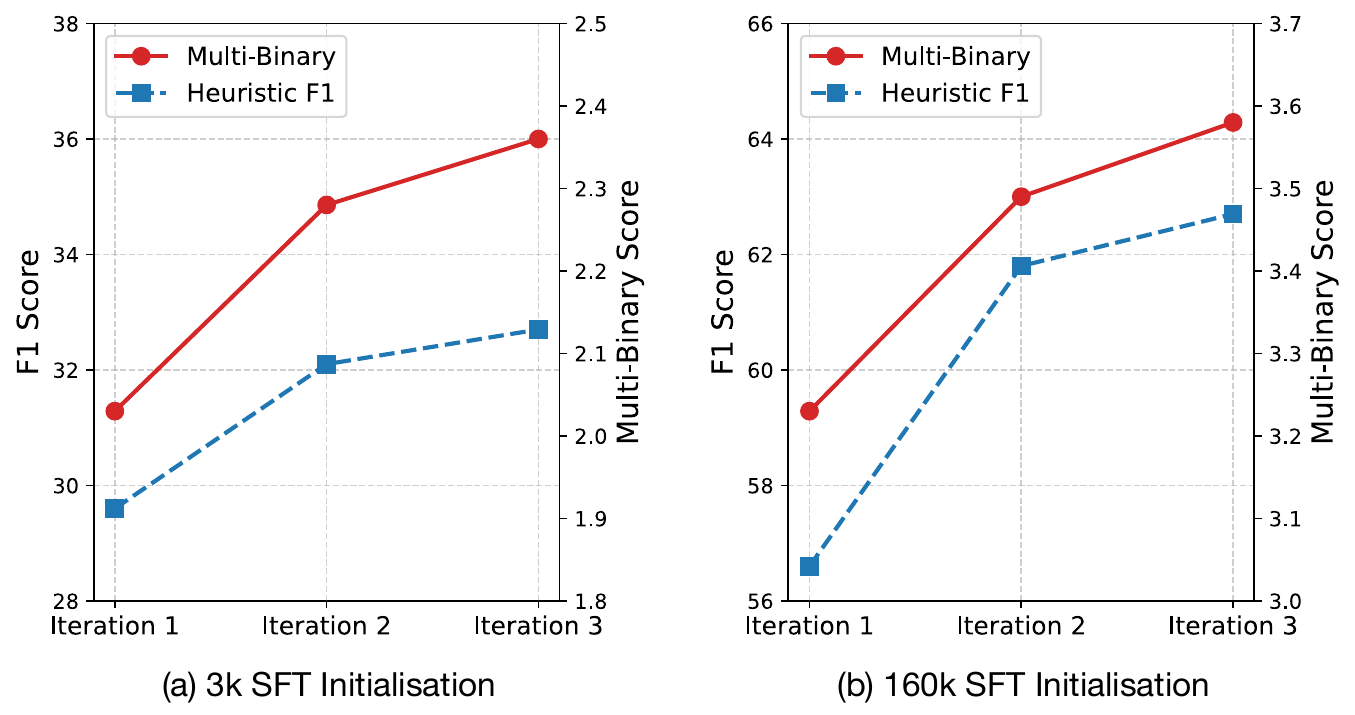}
    \vspace{-0.4cm}
    \caption{Rewarding signals during each iteration of Chart2Code in (1) 3k and (2) 160k SFT initialisation.}
    \vspace{-0.1cm}
    \label{fig:reward_signal_iterations}
\end{figure}

The iteration-wise results in Table~\ref{table:experimental_result} demonstrate that our offline iterative preference learning strategy enables models to achieve progressively higher performance, yielding substantial improvements in both code generation and visual fidelity. This trend of refinement is also reflected in the reward signals observed across iterations. As shown in Figure~\ref{fig:reward_signal_iterations}, dual-modality reward scores exhibit consistent upward trajectories under both the 3k and 160k initialization settings, indicating steady and effective model alignment over time. Furthermore, Chart2Code delivers improvements across all evaluation dimensions, with particularly notable gains in layout, text content, and chart type accuracy. The framework also yields positive performance gains across all difficulty levels in the ChartMimic benchmark, with the most pronounced improvements observed on medium-difficulty samples. These findings are supported with further evidence in the supplementary material.

\subsection{Ablation Study}
We perform extensive ablation studies to assess the contribution of individual components within the Chart2Code framework across three base MLLMs. Owing to computational resource constraints, our ablation experiments are conducted under the 3k supervised fine-tuning initialization setting and evaluated on ChartMimic.

\vspace{-0.1cm}
\subsubsection{Model-agnostic Generalization}
We validate the generalizability of our Chart2Code framework across three distinct MLLMs, including LLaVA-v1.6-7B, InternVL2.5-2B, and Qwen2-VL-7B, as shown in Table~\ref{table:ablation_model_reward}. Despite their differing architectures and capacities, all three models consistently benefit from the dual preference-guided refinement strategy, showing sustained gains in code correctness and visual alignment under all metrics. These results show that Chart2Code is not only effective but also model-agnostic—serving as a plug-and-play training framework capable of enhancing chart-to-code generation across a diverse range of MLLMs.

\begin{figure*}
    \centering
    \includegraphics[width=\linewidth]{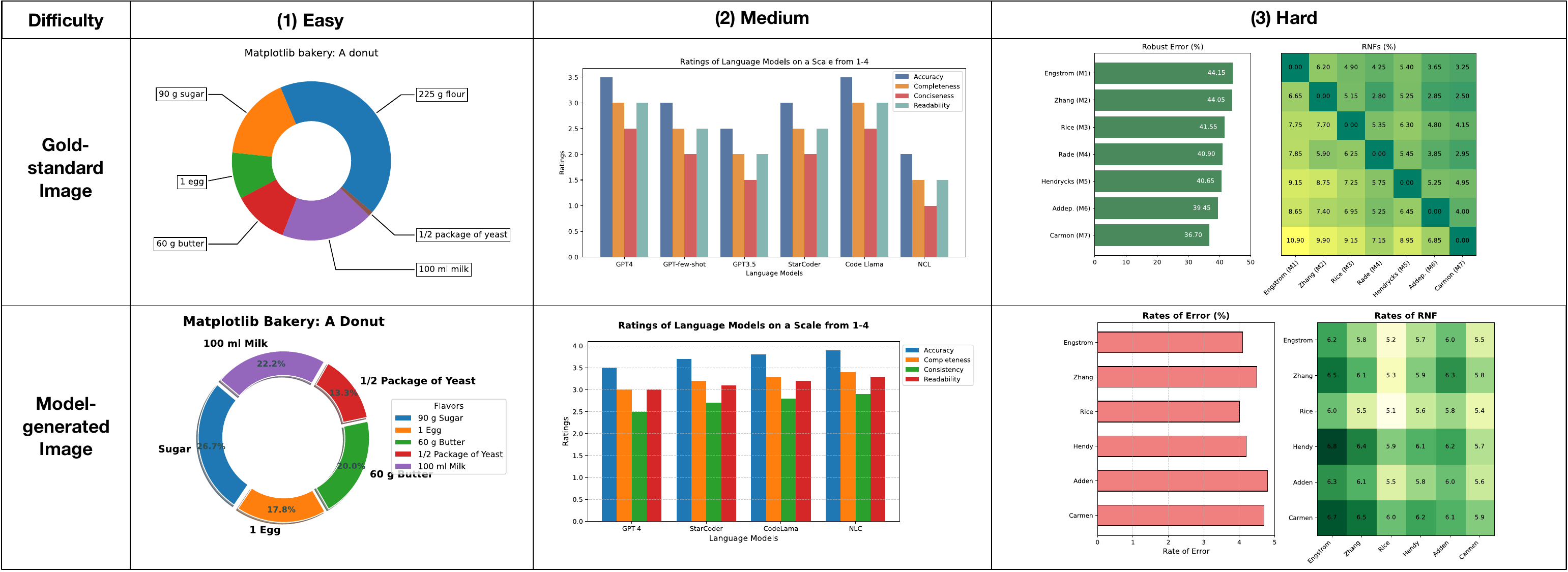}
    \caption{Case study of chart-to-code generation with Chart2Code framework. The three cases are chosen from the three difficulty levels in ChartMimic respectively.}
    \label{fig:case_study}
\end{figure*}

\begin{table}[t]
    \centering
    \renewcommand*{\arraystretch}{1.1}
    \setlength\tabcolsep{2pt}
    \caption{Ablation study of base MLLMs and rewarding signal on ChartMimic. }\vspace{-0.1cm}
    \label{table:ablation_model_reward}
    \scalebox{0.8}{
    \begin{tabular}{p{0.2cm}p{2.2cm}p{1.5cm}<{\centering}p{1.5cm}<{\centering}p{1.8cm}<{\centering}p{2.0cm}<{\centering}}
        \toprule
         \multicolumn{2}{c}{\textbf{Model}}  & \textbf{Exec. Rate} &  \textbf{Heuri. F1} &  \textbf{GPT Conti.} & \textbf{Multi-Binary} \\
        \midrule
        \multicolumn{2}{l}{InternVL2.5-2B} & 48.8  & 21.9  & 22.6  & 1.31 \\
        & \textit{Initial SFT}  & 34.2  & 20.3  & 19.8  & 1.43 \\
        & + Heuristic F1 & 48.6  & 31.4  & 28.2  & 1.41 \\
        & + GPT Conti. & \textbf{56.8}  & 31.3  & 29.2  & 1.59 \\
        & + Multi-Binary  & 52.2  & 31.7  & 29.9  & 1.61 \\
        & + Dual Scoring & 53.1 & \textbf{32.7} & \textbf{31.4} & \textbf{1.66} \\
        \hline
        \multicolumn{2}{l}{LLaVA-v1.6-7B} & 55.6 & 23.6  & 20.2 & 1.09 \\
        & \textit{Initial SFT} & 56.2  & 24.8  & 23.2  & 1.38 \\
        & + Heuristic F1 & 62.2  & 27.0  & 24.8  & 1.41 \\
        & + GPT Conti. & 62.0  & 25.9  & 24.0  & 1.38 \\
        & + Multi-Binary & \textbf{68.0}  & 26.7  & 25.2 & 1.48 \\
        & + Dual Scoring & 63.2 & \textbf{27.2} & \textbf{25.8} & \textbf{1.52} \\
        \hline
        \multicolumn{2}{l}{Qwen2-VL-7B} & 62.2 & 30.0 & 28.9  & 1.09 \\
        & \textit{Initial SFT} & 57.6  & 41.0 & 30.6  & 1.28\\
        & + Heuristic F1 & 60.6  & 41.5  & 31.5  & 1.19 \\
        & + GPT Conti. & 59.6 & 40.9 & 31.4  & 1.20 \\
        & + Multi-Binary & \textbf{62.8} & 42.5  & 32.4  & 1.35 \\
        & + Dual Scoring & 62.1 & \textbf{42.9} & \textbf{33.3} & \textbf{1.36} \\
        \bottomrule
    \end{tabular}
    }
\end{table}

\subsubsection{Role of Dual Rewarding} 
We assess the effectiveness of the dual rewarding mechanism in Chart2Code by comparing it against single-modality reward configurations. To directly evaluate the quality of the reward signals, we measure the agreement between each scoring method and the gold-standard preferences on the feedback evaluation set (Section~\ref{section:ref_dataset_details}), quantified by the proportion of preference pairs for which the scoring method selects the same winner as the ground truth. As shown in Table~\ref{table:evaluator_result}, the proposed dual scoring approach achieves the highest accuracy at 99.8\%, followed by our multi-dimensional binary scoring method at 96.5\%. While dual scoring yields a smaller set of valid preference pairs due to its stricter selection criteria, it consistently leads to stronger downstream performance across all three MLLMs on the ChartMimic benchmark (Table~\ref{table:ablation_model_reward}). 
Additionally, the proposed multi-dimensional binary scoring method outperforms GPT-4o-based scoring in both reward accuracy and downstream model performance across all evaluated MLLMs. This highlights the advantage of incorporating explicit, aspect-level reasoning in the feedback generation process, which produces more reliable and informative reward signals for guiding preference learning.

\begin{table}
\renewcommand*{\arraystretch}{1.1}
\caption{Rewarding accuracy (left) and drop rate (right) during preference construction under different reward signals.}
\vspace{-0.1cm}
\label{table:evaluator_result}
\scalebox{0.9}{
\begin{tabular}{lccc}
\toprule
\textbf{Reward Signal} & \textbf{Prop. w. Corr. Winner} &  \textbf{Prop. of Dropping} \\ 
\midrule
Heuristic F1 & 94.4 & 91.2 \\
GPT Conti. & 91.2 & 96.4 \\ 
Multi-Binary & 96.5 & 94.4 \\ 
Dual Scoring & 99.8 & 85.7 \\
\bottomrule
\end{tabular}}
\end{table}

\subsubsection{Role of Preference Construction} 
Our iterative training method relies on the preference construction using model-generated codes and synthetic variants, and the preference learning algorithm. To assess each component’s impact, we conduct an ablation study with three settings: (1) \textit{SFT on Gold} - supervied finetuning on all gold examples; (2) \textit{PL on Variants} - preference learning on pairs of synthetic variants; and (3) \textit{PL on (Gold, Resp.)} - preference learning on pairs of gold-standard codes and model-generated codes. The latter two share the same initialization as our method. As shown in Table \ref{table:ablation_preference_learning}, our method consistently leads to superior performance. The inclusion of model-generated codes enables the model to iteratively refine its outputs, while synthetic variants serve as a structured reference that bridges the gap between gold-standard and self-generated codes. This hybrid approach shows more effective than relying solely on gold-standard examples, fostering better adaptation and improved generalization.

\begin{table}[!htp]
    \centering
    \renewcommand*{\arraystretch}{1.1}
    \setlength\tabcolsep{2pt}
    \caption{Ablation study of preference learning settings.}\vspace{-0.1cm}
\label{table:ablation_preference_learning}
    \scalebox{0.8}{
    \begin{tabular}{p{0.2cm}p{2.5cm}p{1.5cm}<{\centering}p{1.5cm}<{\centering}p{1.8cm}<{\centering}p{2.0cm}<{\centering}}
        \toprule
         \multicolumn{2}{c}{\textbf{Model}}  & \textbf{Exec. Rate} &  \textbf{Heuri. F1} &  \textbf{GPT Conti.} & \textbf{Multi-Binary} \\
        \midrule
        \multicolumn{2}{l}{LLaVA-v1.6-7B} & 55.6 & 23.6  & 20.2 &  1.09 \\
        & SFT on Gold  &  56.2  & 27.1 &  24.2  & 1.05 \\
        & PL on Variants  & 52.4  & 19.3  & 17.6  & 0.60\\
        & PL on (Gold, Resp.)  & 49.6 & 24.9  & 22.8  & 0.91\\
        & Chart2Code &  \textbf{63.2} &  \textbf{27.2} &  \textbf{25.8} &  \textbf{1.52} \\
        \hline
    \end{tabular}
    }
\end{table}

\section{Case Study}
To qualitatively assess the MLLM's chart-to-code generation capability guided under our framework, we conduct the case study in Figure~\ref{fig:case_study} showcasing three representative examples generated by LLaVA-v1.6-7B trained under the Chart2Code framework with a 3k SFT initialization. These examples are drawn from the easy, medium, and hard difficulty levels in the ChartMimic, respectively. For the easy-level donut pie chart, the model correctly identifies the chart type, color scheme, and textual elements, with only a minor stylistic deviation in legend usage. In the medium-level grouped bar chart, it preserves the overall layout and textual structure, despite omitting two data groups and mismatching some color-label associations. For the hard-level multi-panel chart, the model effectively captures the complex layout and structure, though minor inaccuracies appear in the heatmap’s data and color mapping.

\section{Conclusion}
We presented Chart2Code, a dual preference-guided refinement framework that addresses the key challenges of chart-to-code generation—namely, the under-constrained nature of the task and the need for multi-dimensional fidelity. By combining dual-modality reward signals with structured variant generation and aspect-aware visual evaluation, our method enables scalable, fine-grained preference learning through offline reinforcement. Experiments across multiple MLLMs and benchmarks show that Chart2Code consistently improves execution accuracy and visual alignment, closing the gap between open-source and proprietary systems. Beyond these empirical gains, our work highlights the broader potential of preference-based learning in multimodal settings, especially for tasks where correctness spans both symbolic and perceptual dimensions. We believe this approach offers a generalizable path forward for aligning MLLMs with structured generation tasks.

\section*{Acknowledgments}
This research is supported by the National Research Foundation, Singapore under its National Large Language Models Funding Initiative (AISG Award No: AISG-NMLP-2024-002). This research is also supported by the Ministry of Education, Singapore, under its AcRF Tier 2 Funding (Proposal ID: T2EP20123-0052). Any opinions, findings, conclusions, or recommendations expressed in this material are those of the author(s) and do not reflect the views of the National Research Foundation or the Ministry of Education, Singapore.

\bibliographystyle{ACM-Reference-Format}
\balance
\bibliography{acmart}

\end{document}